\documentclass[11pt]{article}

\usepackage[letterpaper,margin=1in]{geometry}

\usepackage[T1]{fontenc}
\usepackage{microtype}

\usepackage{newtxtext}
\usepackage{newtxmath}

\usepackage{authblk}

\usepackage{graphicx}
\usepackage{tabularx}
\usepackage{multirow}
\usepackage{makecell}
\usepackage{booktabs}
\usepackage{array}

\newcolumntype{C}{>{\centering\arraybackslash}X}

\usepackage{amsmath}

\usepackage{xurl}
\usepackage[hidelinks]{hyperref}
\usepackage[nameinlink,noabbrev]{cleveref}

\usepackage{caption}

\usepackage[round,authoryear]{natbib}

\title{
\textbf{Det-LIME: Detector-Aware, Multi-Instance Local Interpretable
Model-Agnostic Explanations for Automated Marine Mammal Detection}
}

\author[1]{Jiayi Zhou}
\author[2]{David W. Johnston}
\author[1]{Brinnae Bent\thanks{
Corresponding author:
\href{mailto:brinnae.bent@duke.edu}{brinnae.bent@duke.edu}
}}

\affil[1]{
Pratt School of Engineering,
Duke University,
Durham,
North Carolina, United States
}

\affil[2]{
Division of Marine Science and Conservation,
Nicholas School of the Environment,
Duke University,
Durham,
North Carolina, United States
}

\date{}

\begin{document}

\maketitle


\begin{abstract}
Despite the rapid uptake of black-box object detectors in marine mammal
research and monitoring, explainability techniques are rarely integrated
into conservation workflows. Furthermore, most classification-oriented
explainability tools are ill-suited to detection tasks involving imagery
of social organisms or those with colonial life histories, as they ignore
multiple detections within a scene and produce single-instance outputs
that blur evidence across individuals. These methods also generate
low-resolution, often biologically irrelevant visuals, limiting their
utility for debugging, targeted data augmentation, and refined data
collection.

We proposed Det-LIME, a detector-aware, multi-instance adaptation of Local
Interpretable Model-Agnostic Explanations (LIME) that produced
instance-specific, box-aligned explanations by combining per-detection
weighting, a proximity kernel that emphasizes regions near each box, and
Intersection-over-Union-based matching to track the same instance across
perturbations. We evaluated Det-LIME on aerial drone imagery for harbor
seal detection, with an additional seabird case study to assess
generality, and compared it with vanilla LIME, Stabilized LIME,
Deterministic LIME, and gradient-based attribution methods.

Using the Attribution Ratio and Max Saliency Hit Rate metrics, we showed
that Det-LIME consistently improved multi-instance attribution. In
practice, these higher-resolution, instance-aware explanations provide
insight into model outputs and support post-processing, debugging, and
actionable improvements in modeling and data collection or augmentation.
\end{abstract}



\section{Introduction}
\label{sec:introduction}

Computer vision and artificial intelligence methods now underpin many ecological workflows: from tracking wildlife populations and assessing habitat change to informing management actions and protected-species policy \citep{gray2022drones,ruzicka2023semantic}. Classification and object detection methods are commonly used within computer vision workflows to automatically identify, count, and locate species and landscape features in large image and video datasets, enabling scalable biodiversity monitoring and informed conservation decisions \citep{osti_1614652, choinski2021stepautomatedspeciesrecognition}. 

As detectors mature, accuracy is no longer the only limiting factor. Practitioners need evidence that predictions are grounded in animal morphology (not background shortcuts), clear views of failure modes, and defensible rationale when setting thresholds that carry operational and ethical costs \citep{gevaert2022explainable,buchelt2024exploring}. This is particularly true for what people refer to as "black-box" models like neural networks with opaque internal "reasoning". Although researchers who develop and train their own models may have access to the model architecture, this level of access is not always available in applied ecological workflows. Models may be developed by external collaborators, incorporated into third-party software, accessed through hosted services, or deployed in monitoring systems that provide predictions without exposing internal feature maps or gradients. Even when the model architecture is available, explanation methods that rely on internal representations generally require architecture-specific choices, including identifying appropriate feature layers and linking the prediction of interest to the corresponding activations and gradients. These requirements can vary across detector architectures and make the same explanation procedure more difficult to apply consistently across models. Despite these needs, explainability is rarely integrated into ecological research or conservation workflows.

Explainability helps provide insights into model predictions by revealing which input features and internal representations drive a model’s decision-making process. One such explainable AI (XAI) technique is Local Interpretable Model-Agnostic Explanations (LIME) \citep{ribeiro2016whyitrustyoulime}.  LIME generates perturbed versions of the input image and observes how the black-box model’s predictions change. It then fits a simple linear model to these perturbations to approximate the local decision boundary, using the linear coefficients to estimate the contribution of each image region. This provides a locally interpretable explanation of the model’s prediction. In the context of marine mammal ecology, LIME is particularly well-suited because it is model agnostic, local, and produces spatially grounded, human-understandable overlays. Because it operates from model inputs and outputs, it does not require users to select internal feature layers or adapt the explanation procedure to specific internal representations. This provides a consistent way to examine predictions across detector architectures, including settings where internal model information is unavailable or difficult to use directly for explanation. These properties align with how ecologists audit individual predictions and allow visual verification that predictions are based on animal morphology rather than background features. LIME’s portability, locality, and communicability make it well-matched to ecological decision-making.

Classification models produce a single, image-level prediction, and explanation methods in this setting attribute evidence to that overall decision. In contrast, multi-object detection systems often produce multiple localized predictions within a single image, where each prediction is defined by a bounding box, a class label, and a confidence score. This distinction is important in ecological images, where multiple animals often appear together, crowded within a single frame. Conventional LIME approaches are designed for per-image classification, not instance-level detection in complex and often unpredictable ecological scenes. In practice, conventional LIME techniques explain a label’s image-level score rather than a detector’s per-box prediction; their perturbations are often coarse, engulfing multiple small animals and blurring evidence across neighboring boxes; they have no mechanism to remain focused on predicted objects as perturbations alter proposals; and they weight these perturbations globally, rather than centering on the queried box. The result is blocky heatmaps that attribute background textures more than morphology in predictions, offering limited guidance for threshold setting, post-processing, or targeted data collection. Explanation methods for multi-object detection in ecological imagery must provide instance-level, spatially precise attributions that are stable for a given detection while accounting for nearby objects and complex background patterns.

To address these limitations, we introduce Det-LIME, a detector-aware, multi-instance adaptation of LIME designed for object detection in ecological imagery. Det-LIME produces instance-specific, box-aligned explanations by fitting a model that approximates the detector's behavior near a target detection. The framework  (i) handles multiple detections through instance weighting, (ii) emphasizes regions near each box with a proximity kernel, and (iii) tracks the same instance under perturbations using intersection over union (IoU)-based matching. The goal was actionable clarity: higher-resolution, per-animal evidence that aligns with how ecologists review detections and ultimately make policy-relevant decisions.

\section{Related Work}\label{sec2}
\subsection{Explainability in Ecology}
Explainable AI (XAI) has shown substantial impact across healthcare and autonomous driving domains, where it reveals diagnostically relevant regions in medical images \citep{brima2024saliency} or justifies robust road detection \citep{mankodiya2022odxai}. In ecology, however, XAI remains comparatively underdeveloped \citep{saarela2024recent}. Recent reviews have discussed the need for explainability \citep{gevaert2022explainable, buchelt2024exploring}, especially in high-stakes conservation contexts; however, most ecological workflows, especially those involving computer vision tasks, employ “black box” models without any integrated explainability.

In classification and regression tasks, XAI methods provide insight into which features or regions drive model predictions. For instance, Local Interpretable Model-agnostic Explanations (LIME) has been used to provide localized insight in species distribution modeling \citep{ryo2021xai} and to highlight influential regions for bird image classification \citep{bird2025lime}, while SHapley Additive exPlanations (SHAP) has been applied to identify environmental drivers of vegetation health \citep{britton2024ndvi}. These approaches help researchers understand feature importance at the image or data level.

For object detection, techniques such as Grad-CAM \citep{9897350} and saliency maps \citep{petsiuk2021blackboxexplanationobjectdetectors} have been adapted to reveal regions associated with individual detections. However, these gradient-based approaches typically require access to the model’s internal architecture or feature maps, limiting their generality. Model-agnostic methods like conventional LIME have been previously applied to object detection but generally explain only a single detected instance per image \citep{Sejr2021SODEx}. Detecting multiple small species in ecological images is particularly challenging and is often required for monitoring marine mammal species that exhibit social clustering in either aquatic or terrestrial habitats. Current methods cannot provide instance-level, spatially precise explanations across multiple detections without access to model internals, creating a need for a model-agnostic approach that supports interpretable and reliable ecological object detection for monitoring and conservation.

\subsection{Local Interpretable Model-Agnostic Explanations (LIME)}
Explainability methods such as LIME provide insight into a model’s behavior for a single prediction of a black-box model \citep{ribeiro2016whyitrustyoulime}. LIME generates perturbed versions of the input image and observes how the model’s predictions change across these perturbations. It then fits a simple interpretable model, typically linear regression, to approximate the local decision boundary and identify the image regions that most influence the prediction.  

However, its reliance on random sampling can produce unstable explanations, prompting the development of variants such as Deterministic LIME (DLIME) and Stabilized LIME (S-LIME) that address these instabilities \citep{zafar2019dlimedeterministiclocalinterpretable, Zhou_2021}. DLIME uses a deterministic hierarchical approach to generate perturbations, while S-LIME aggregates the results of multiple runs to improve consistency. These adaptations make local explanations more reliable.

Despite these advances, LIME and its variants are designed for classification tasks, where a single prediction is produced for the entire image. Multi-object detection presents a different challenge, particularly in ecological applications, because the model produces multiple predictions, each corresponding to a distinct object. Traditional LIME cannot provide instance-level explanations in this setting.

We addressed this limitation by developing Det-LIME, a detector-aware adaptation of LIME that extends local explanations to multi-object detection. This approach preserves the model-agnostic strengths of LIME while providing instance-level, spatially specific explanations for ecological imagery.

\section{Methods}\label{sec3}
\subsection{Det-LIME: Enhanced Multi-Instance LIME for Object Detection}
\begin{figure*}[htbp]
    \centering
    \includegraphics[width=1\textwidth]{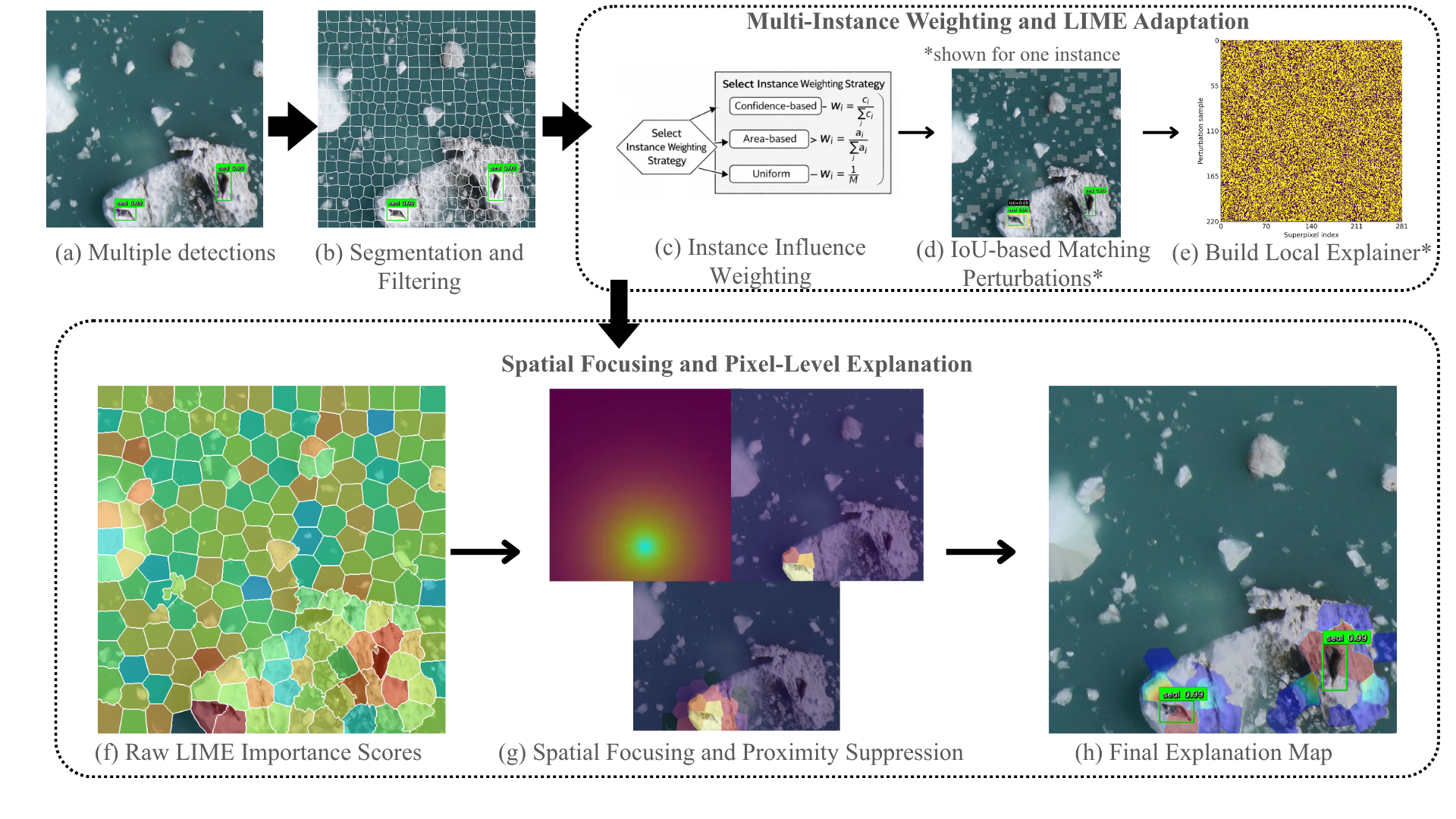}
    \caption{An illustration of the Det-LIME explanation process, applied to the output of a Faster R-CNN model detecting seals. This diagram illustrates our method, starting from an image with multiple detections (a) that was partitioned into superpixels (b). After weighting each detected instance (c), the system scored numerous perturbed samples (d) and used them to fit a model that approximated the detector's behavior near a target detection (e). This yielded raw importance scores (f) for each superpixel, which were then spatially focused (g) to produce the final, box-aligned relevance map (h) that highlighted the key visual evidence for each detection.}
    \label{fig:detlime_pipeline}
\end{figure*}model-agnostic

We extended the Local Interpretable Model-agnostic Explanations (LIME) framework to multi-instance object detection in imagery generated during drone surveys of harbor seals in glacial fjords in Glacier Bay National Park and Preserve, \citep{aghakishiyeva2025photorealistic}, alongside a complementary case focused on drone surveys of colonial seabirds \citep{hayes2021drones} (details on both applications below). Our approach introduced three key enhancements: an improved segmentation and filtering pipeline, a flexible multi-instance weighting mechanism, and a refined spatial focusing strategy. These refinements generated explanations that were more detailed and better aligned with the objects detected by the model (Figure 1h). The resulting explanation maps highlighted image regions that most influenced the model's detection decision, which could occur both within and around the predicted bounding box depending on the visual features and contextual cues used by the detector.

\subsubsection{Segmentation and Filtering}

The first enhancement refined image segmentation. Given an input image, we partitioned it into superpixels using an enhanced Simple Linear Iterative Clustering (SLIC) algorithm with additional hyperparameters for increased flexibility. Each superpixel satisfied constraints on compactness, minimum size, and connectivity, and the superpixels collectively covered the image without overlap. An optional filtering step removed visually insignificant segments, such as regions that were uniformly dark. This reduced background noise and improved the fidelity of the final explanation.

\subsubsection{Multi-Instance Weighting and LIME Adaptation}

LIME is designed to explain a single prediction, while multi-object detection models typically return multiple detections of the same class within a single image. To bridge this gap, we extended LIME to a multi-instance setting by explicitly defining target instance selection, instance weighting, perturbation scoring, and linear explanation fitting.

\paragraph{Target Instance Identification and Weighting}

Given an input image, the object detection model produces a set of detections
\[
R = \{(b_j, c_j, p_j)\}, \quad j = 1, \dots, M,
\]
where \(b_j\) denotes the \(j\)-th predicted bounding box, \(c_j\) the associated class label, \(p_j\) the detection confidence, and \(M\) the total number of detections. For a target class \(c^*\) and confidence threshold \(\theta\), we define the set of target instances as
\[
\mathcal{I} = \{ b_j \in R \mid c_j = c^*, \; p_j \geq \theta \}.
\]
This set contains all sufficiently confident detections belonging to the class of interest.

Each target instance \(b_i \in \mathcal{I}\) was assigned a weight \(w_i\) that determined its contribution to the explanation. We considered three weighting strategies:
\[
w_i =
\begin{cases}
\dfrac{p_i}{\sum_{k=1}^{K} p_k}, & \text{Confidence-based weighting} \\[1.5ex]
\dfrac{\mathrm{area}(b_i)}{\sum_{k=1}^{K} \mathrm{area}(b_k)}, & \text{Area-based weighting} \\[0.5ex]
\dfrac{1}{K}, & \text{Uniform weighting}
\end{cases}
\]
Here, \(K = |\mathcal{I}|\) denotes the number of target instances. These strategies allowed the explanation to emphasize detections according to model confidence, spatial extent, or equal contribution.

\paragraph{Multi-Instance Perturbation and Scoring}

Following the LIME framework, we generated a set of perturbed images by masking different subsets of superpixels. Let \(I'_m\) denote the \(m\)-th perturbed image. Applying the detector to \(I'_m\) yielded a corresponding set of detections
\[
R_m = \{(b_j, c_j, p_j)\},
\]
defined analogously to \(R\).

For each target instance \(b_i \in \mathcal{I}\), we computed an instance-level score \(s_i^{(m)}\) that measured how well the instance was preserved under the perturbation:
\[
s_i^{(m)} = w_i \cdot \max_{\substack{b_j \in R_m \\ c_j = c^* \\ \mathrm{IoU}(b_i, b_j) > \tau}}
\Big( \min(p_j, 0.95) \cdot (0.7 + 0.3 \cdot \mathrm{IoU}(b_i, b_j)) \Big).
\]
Here, \(\mathrm{IoU}(b_i, b_j)\) quantifies the spatial overlap between the original instance \(b_i\) and a detected bounding box \(b_j\) in the perturbed image:
\[
\mathrm{IoU}(b_i, b_j) = \frac{\mathrm{area}(b_i \cap b_j)}{\mathrm{area}(b_i \cup b_j)}.
\]
The overlap threshold \(\tau\) filtered out detections with minimal spatial correspondence to the original instance, ensuring that only perturbed detections that could be meaningfully matched back to the queried object contributed to the attribution score. For matched detections, confidence values above 0.95 were capped at 0.95 for scoring rather than removed from the analysis. Thus, for example, a perturbed detection with confidence 0.98 contributed a confidence value of 0.95 to the instance-level score. Detection confidence was capped at 0.95 to prevent a single highly confident prediction from dominating the multi-instance aggregation. This cap helped keep instance contributions comparable across detections within the same image and allowed the final attribution map to remain sensitive to spatial perturbations rather than being driven by the detector’s confidence scale. The overlap-dependent term assigned higher scores to detections that better aligned with the original instance. The confidence cap of 0.95 and the IoU weighting values of 0.7 and 0.3 were specific to our multi-instance extension and were not inherited from the original LIME formulation. These values were selected as reasonable default settings based on initial pilot runs and stability considerations.

The overall score for the perturbed image was obtained by aggregating contributions from all target instances:
\[
S^{(m)} = \sum_{i \in \mathcal{I}} s_i^{(m)}.
\]
This scalar score reflected the extent to which the set of target detections was preserved under the perturbation and served as the response variable for LIME.

\paragraph{LIME Linear Model Fitting}

Superpixel importance was estimated by fitting a weighted linear model following the standard LIME formulation:
\[
g(z) = \beta_0 + \sum_{j=1}^{N} \beta_j z_j, \quad z_j \in \{0,1\},
\]
where \(N\) is the number of superpixels, \(z_j = 1\) indicates that superpixel \(S_j\) was present in the perturbed image, and \(z_j = 0\) indicates that it was masked. The coefficient \(\beta_j\) represents the contribution of superpixel \(S_j\) to preserving the aggregated multi-instance score \(S^{(m)}\), thereby capturing its importance across all target detections. A linear surrogate was used because each coefficient can be directly associated with an individual superpixel, allowing its contribution to the prediction to be mapped back to a specific image region. The surrogate approximates the detector only within the local set of perturbations and does not assume that the underlying detector itself is linear. 

\subsubsection{Spatial Focusing and Pixel-Level Explanation}

We introduced a spatial focusing mechanism that translated superpixel-level LIME scores into a refined, pixel-level saliency map. For each pixel \((x, y)\), two key factors were considered for each target instance \(b_i \in \mathcal{I}\):  

\begin{enumerate}
    \item \textbf{Superpixel overlap with the instance:} Each superpixel \(S_j\) may cover part of a target instance. The fraction of the superpixel overlapping with \(b_i\) was computed as
  
    \[
    \text{overlap}(S_j, b_i) = \frac{|S_j \cap b_i|}{|S_j|}.
    \]  
    This term ensured that pixels within superpixels that covered the object received higher importance.

    \item \textbf{Spatial proximity to the instance center:} Pixels closer to the center of a bounding box were assumed to be more informative. We defined a distance-based weight 
    \[
    \text{distance\_weight}(x,y; b_i) = \exp\Bigg(- \frac{d((x,y), \text{center}(b_i))}{r_i / 3}\Bigg),
    \]  
    where \(d(\cdot,\cdot)\) is the Euclidean distance and \(r_i\) is the diagonal length of \(b_i\). This factor gradually decreased the importance of pixels farther from the object center, emphasizing central regions while still assigning partial credit to surrounding pixels.
\end{enumerate}

The final pixel-level explanation combined these factors with the superpixel importance \(\beta_j\) from the LIME model and the instance weight \(w_i\) to form a unified multi-instance saliency map:  
\[
E(x,y) = \max_{i \in \mathcal{I}} \; \max_{S_j \notin B} 
\Big( \beta_j \cdot w_i \cdot \text{overlap}(S_j, b_i) \cdot \text{distance\_weight}(x,y; b_i) \Big),
\]  
where \(B\) denotes excluded black or filtered segments. The inner maximization selected the superpixel most relevant for each instance, while the outer maximization ensured that each pixel reflected the contribution of the most influential instance among all targets. Finally, the explanation map \(E\) was normalized to \([0,1]\) and could be optionally smoothed with a Gaussian filter. 

This approach allowed the explanation to accurately highlight pixels that contributed to detecting multiple instances, accounting for both the superpixel-level importance and the spatial structure of the objects.

\subsection{Evaluation Protocol}
We assessed the performance of four local explanation methods to evaluate the effectiveness of the proposed approach: LIME \citep{ribeiro2016whyitrustyoulime}, its stabilized variants DLIME \citep{zafar2019dlimedeterministiclocalinterpretable} and SLIME \citep{Zhou_2021}, and the proposed Det-LIME. LIME served as the baseline, as Det-LIME was built directly on its framework. DLIME and SLIME were included because they address LIME’s instability issues by providing more consistent explanations. Unlike the other methods, which generate explanations only for the single highest-confidence object in an image, Det-LIME generated a comprehensive attribution map for all detected objects. For this analysis, Det-LIME was configured in “Uniform” mode, assigning equal importance to every detected object. Under this configuration, each detected instance contributed equally to the aggregated explanation, independent of its confidence score or bounding-box size, thereby establishing a standardized baseline for multi-instance aggregation. While alternative strategies, such as confidence-based or area-based weighting, can be employed to emphasize detection reliability or object scale, respectively, they introduce additional dependencies on detector-specific outputs. In this work, uniform weighting was adopted to maintain a consistent, assumption-light aggregation framework across diverse instances.

Performance was assessed using two commonly used metrics that measure how well attributions match ground-truth object locations. These metrics were used to assess how well each explanation aligned with the detected object region, rather than as standalone measures of detector quality. Because Det-LIME explains detector outputs, these metrics necessarily depend on the detector’s predicted bounding boxes, class labels, and confidence scores. To reduce the influence of low-confidence detections, Det-LIME explanations were evaluated for detections above a confidence threshold of 0.5. In addition, explanation methods were compared using the same trained detector outputs, datasets, and detection targets, thereby reducing variation attributable to different model predictions or evaluation settings. With the detector output held constant across methods, differences in the resulting scores were more related to attribution behavior than to detection performance.

Let $A \in \mathbb{R}^{H \times W}$ be the attribution map for a detected instance, normalized to $[0,1]$, and let $B \subset \{1,\dots,H\} \times \{1,\dots,W\}$ denote the ground-truth bounding box for that instance. The first metric, \textbf{Attribution Ratio (AR)}, measures the proportion of important pixels that fall inside the object region, a metric widely adopted in prior work \citep{zhou2015learningdeepfeaturesdiscriminative, Selvaraju_2019}. We thresholded the attribution map to keep only mid-to-high importance pixels:
\[
\mathcal{P}_\tau = \{(i,j) \mid A_{ij} \ge \tau \},
\]
and computed
\[
\text{AR} = \frac{|\mathcal{P}_\tau \cap B|}{|\mathcal{P}_\tau|}.
\]
We used a threshold of $\tau = 0.3$ to remove low-intensity pixels in the attribution map. This threshold provided a conservative cutoff that improved the visual focus of the attention map by removing diffuse, low-signal pixels while retaining high-importance structures associated with the detected objects. This ensured that the metric reflected the alignment between the high-intensity pixels and the object regions. The second metric, \textbf{Max Saliency Hit Rate (MSHR)}, measures whether the single most important pixel lies within the object box, another commonly used evaluation metric \citep{zhang2016topdownneuralattentionexcitation}. Let
\[
(i^*, j^*) = \arg\max_{(i,j)} A_{ij},
\]
then the hit for one instance is
\[
\text{MSHR} = \mathbb{I}\big((i^*, j^*) \in B \big),
\]
where $\mathbb{I}$ is the indicator function. The final score was calculated as the average over all instances.

Together, AR and MSHR captured whether attribution was concentrated within the object region and whether the most salient pixel fell inside the ground-truth box. These metrics were applied across all instances in the held-out test dataset to evaluate the methods’ performance in a realistic, multi-object scenario. For Det-LIME, each of its multiple explanations was matched to its corresponding ground-truth object before calculating the metrics. In parallel, the single explanation from LIME, DLIME, and SLIME was evaluated against every ground-truth object present in the image. These results were then aggregated across the entire dataset for each of the four methods. The Mean Attribution Ratio was obtained by averaging the individual Attribution Ratios across all instances, while the Max Saliency Hit Rate was the overall percentage of instances where the most salient pixel fell within the ground-truth box.

Given the difference between Det-LIME’s multi-object output and the single-object output of other methods, we adopted a second evaluation strategy. Evaluating all objects together can disadvantage single-instance methods in multi-object images, whereas Det-LIME naturally provides attribution maps for all detected objects. To isolate this effect, the specific object selected by LIME, DLIME, and SLIME in each image was first identified. A one-to-one subset was then created by pairing these instances with Det-LIME’s corresponding explanations. Recalculating the metrics on this filtered dataset produced the Adjusted Mean Attribution Ratio and Adjusted Max Saliency Hit Rate, providing a perspective that allowed a fair comparison across methods by controlling for the influence of multiple instances in the evaluation metrics.

For both evaluation strategies, we reported the total number of evaluated detection instances used to compute each metric. An evaluated instance refers to a predicted detection that could be matched to a corresponding ground-truth bounding box and for which an attribution map was included in the metric calculation. To ensure that the evaluation focused on interpretable explanations for valid object detections, unmatched or invalid detections were excluded before aggregating the instance-level metrics. For the all-instance evaluation, the reported number represents all valid matched detections across the held-out test images for each method. For the adjusted evaluation, it represents the one-to-one subset of instances selected by the single-instance LIME-based methods and matched to the corresponding Det-LIME explanations. This distinction is important because the all-instance evaluation reflects each method’s native explanation setting, whereas the adjusted evaluation controls for differences between multi-instance and single-instance explanation outputs.

In addition to quantitative evaluation, we provided a visual comparison of Det-LIME with several existing explainability methods, including the gradient-based class activation mapping approach LayerCAM \citep{Jianglayercam}, LIME \citep{ribeiro2016whyitrustyoulime}, DLIME \citep{zafar2019dlimedeterministiclocalinterpretable}, and SLIME \citep{Zhou_2021}. LayerCAM leverages internal convolutional layers to produce high-resolution attributions, enabling precise localization of salient features. In contrast, Det-LIME is model-agnostic and can be applied to any object detector, generating heatmaps that indicate which image regions contribute to each predicted bounding box. LIME, DLIME, and SLIME provide explanations for only a single instance within an image, whereas Det-LIME produces explanations for all detected objects in the image. These visual comparisons assessed aspects of explanation behavior that are not fully captured by spatial overlap metrics, including compactness, interpretability, and the ability to represent multiple detections within the same image. By visually comparing these methods, we offered a qualitative view of how Det-LIME addressed the challenges of multi-object detection and provided a model-agnostic approach for interpretable explanations.

\subsection{Det-LIME in Ecological Applications}
We applied the above-described Det-LIME workflow to two different architectures trained on two object detection architectures trained on two datasets. We applied Det-LIME to a harbor seal detection task using Faster R-CNN \citep{ren2016fasterrcnnrealtimeobject} (data collection and labeling methods in Appendix A.1; modeling methods in Appendix B.1). As a broader example demonstrating the applicability of the technique beyond marine mammals and across detection models, we additionally evaluated Det-LIME on a penguin detection task using a YOLOv9 model \citep{wang2024yolov9learningwantlearn}, fine-tuned on a seabird dataset containing Southern Rockhopper Penguins among breeding black-browed albatrosses \citep{hayes2021drones} (modeling methods in Appendix B.2). Both examples focused on animals that are found in groups or colonies, and were representative of applications aimed at understanding animal density, abundance, and distributions in relation to environmental factors. Det-LIME was applied only to the held-out test set images in both cases. Code to apply Det-LIME is available here: \url{https://osf.io/d456u/?view_only=98eef0382ba745d9a1e7b89e2cd9b4a9}

\section{Results}
\subsection{Evaluation of Det-LIME}

\begin{table}[!htbp]
\centering
\caption{Comparison of LIME Variants Across Datasets and Detection Models}
\label{tab:lime_results}

\setlength{\tabcolsep}{3pt}
\renewcommand{\arraystretch}{1.15}

{\footnotesize
\begin{tabularx}{\textwidth}{
    @{}
    p{0.20\textwidth}
    l
    C
    C
    C
    C
    @{}
}
\toprule

\textbf{Dataset and Detection Model}
& \textbf{LIME Variant}
& \makecell[c]{\textbf{Mean Attribution}\\\textbf{Ratio}}
& \makecell[c]{\textbf{Max Hit}\\\textbf{Rate}}
& \makecell[c]{\textbf{Adjusted Mean}\\\textbf{Attribution Ratio}}
& \makecell[c]{\textbf{Adjusted Max}\\\textbf{Hit Rate}} \\

\midrule

\multirow{4}{*}{%
    \makecell[l]{Harbor Seal Detection\\\textit{Faster R-CNN}}
}
& LIME
& \makecell[c]{16.23\%\\($\pm$17.85\%)}
& 17.91\%
& \makecell[c]{22.14\%\\($\pm$18.74\%)}
& 25.33\% \\

& SLIME
& \makecell[c]{25.02\%\\($\pm$24.36\%)}
& 14.93\%
& \makecell[c]{36.32\%\\($\pm$24.37\%)}
& 22.67\% \\

& DLIME
& \makecell[c]{24.17\%\\($\pm$24.09\%)}
& 31.34\%
& \makecell[c]{37.26\%\\($\pm$23.79\%)}
& 52.00\% \\

& Det-LIME
& \makecell[c]{\textbf{48.76\%}\\($\pm$29.51\%)}
& \textbf{92.42\%}
& \makecell[c]{\textbf{52.34\%}\\($\pm$29.65\%)}
& \textbf{96.00\%} \\

\midrule

\multirow{4}{*}{%
    \makecell[l]{Penguin Detection\\\textit{YOLOv9}}
}
& LIME
& \makecell[c]{3.67\%\\($\pm$11.46\%)}
& 3.42\%
& \makecell[c]{29.55\%\\($\pm$20.44\%)}
& 27.96\% \\

& SLIME
& \makecell[c]{4.23\%\\($\pm$12.94\%)}
& 3.46\%
& \makecell[c]{35.11\%\\($\pm$20.94\%)}
& 28.67\% \\

& DLIME
& \makecell[c]{4.71\%\\($\pm$14.24\%)}
& 4.38\%
& \makecell[c]{\textbf{40.15}\%\\($\pm$21.41\%)}
& 36.20\% \\

& Det-LIME
& \makecell[c]{\textbf{39.25\%}\\($\pm$28.19\%)}
& \textbf{82.20\%}
& \makecell[c]{39.79\%\\($\pm$29.08\%)}
& \textbf{98.92\%} \\

\bottomrule
\end{tabularx}
}

\vspace{0.15cm}

\begin{minipage}{\textwidth}
\footnotesize
\textit{Note.}
Values were computed over matched detection instances rather than images.
Only valid matched detections with corresponding attribution maps were
included in the instance-level evaluation. In the all-instance evaluation,
the harbor seal dataset included 134 evaluated instances for LIME, SLIME,
and DLIME, and 132 for Det-LIME; the penguin dataset included 2,720 evaluated
instances for LIME, SLIME, and DLIME, and 2,450 for Det-LIME. The adjusted
metrics were computed on a one-to-one matched subset selected by the
single-instance methods, including 75 harbor seal instances and 279 penguin
instances.
\end{minipage}

\end{table}

\begin{figure*}[htbp]
    \centering
    \includegraphics[width=0.9\textwidth]{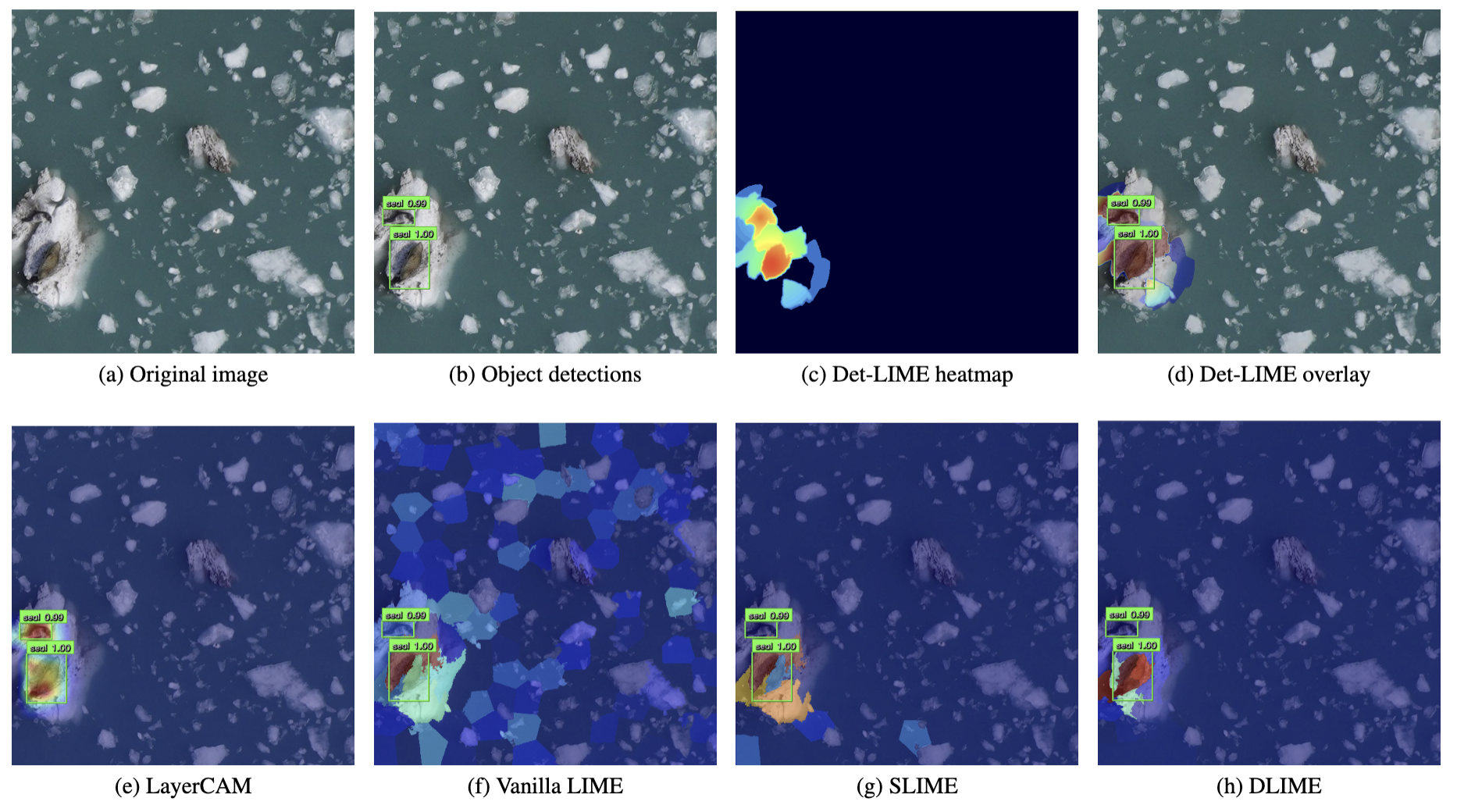}
    \caption{Comparison of different LIME variants for the harbor seal multi-instance detection task using Faster R-CNN. (a) Original image, (b) object detections with bounding boxes, (c) Det-LIME explanation heatmap, (d) Det-LIME overlay, (e) LayerCAM, (f) vanilla LIME, (g) LIME variant SLIME, (h) LIME variant DLIME. In the attribution maps, a color spectrum represents relative importance, with warm colors (e.g., red, orange, yellow) indicating high contribution and cool colors (e.g., blue) indicating little or no contribution.}
    \label{fig:lime_comparison}
\end{figure*}

Our evaluation demonstrated that Det-LIME addressed the major limitations of vanilla and variant LIME approaches in ecological multi-instance detection. The number of evaluated detection instances was reported in Table 1 to clarify the denominator used for Attribution Ratio and Max Saliency Hit Rate, as both metrics were averaged over individual detections rather than over images. As shown in Table 1, Det-LIME substantially improved both the Attribution Ratio and Max Saliency Hit Rate across harbor seal and penguin detection tasks, with gains of more than 20-30 percentage points relative to the strongest baselines in several comparisons. These improvements supported Det-LIME’s ability to provide instance-specific explanations. By aligning perturbations with bounding boxes and performing IoU-based matching between detections in the original and perturbed images, Det-LIME limited attribution drift, preserved the correspondence between saliency and individual objects, and reduced spurious signals in backgrounds. 

In comparison, baseline methods had lower Attribution Ratio and Max Saliency Hit Rate (Table 1), reflecting attributions that blurred across multiple objects, overemphasized background correlations, and lacked spatial precision, making it difficult to identify the features truly driving each detection. Visual comparisons in Figure 2 reinforced this quantitative evidence. While vanilla LIME and its stabilized variants produced blocky, cluttered maps that often highlighted ice, rock, or water textures, Det-LIME yielded sharper, more coherent attributions that corresponded more closely to annotated animal regions. This shift was especially pronounced in the penguin detection setting, where small, densely clustered animals in colonies posed a severe challenge for coarse perturbation methods; Det-LIME’s high Max Saliency Hit Rate demonstrated its robustness in precisely these contexts. Together, these results showed that Det-LIME adapted LIME’s perturbation framework to detection “in the wild,” producing cleaner, more actionable explanations that better supported threshold setting, post-processing, and targeted ecological data collection.

\section{Discussion}

\begin{figure*}[htbp]
    \centering
    \includegraphics[width=0.9\textwidth]{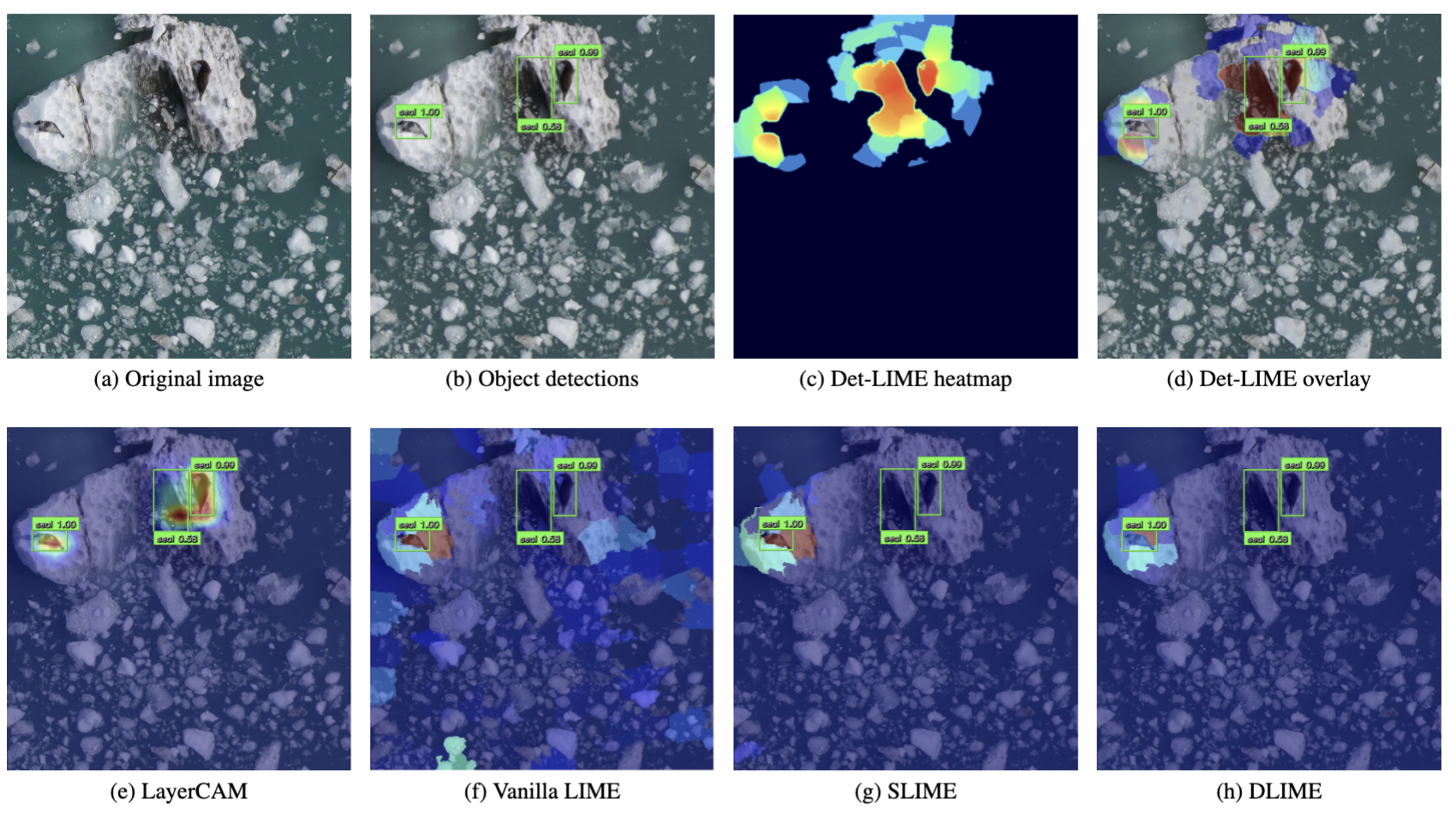}
    \caption{Visual explanations of a failure case where black ice was misidentified as a seal in the harbor seal multi-instance detection task using Faster R-CNN. (a) Original image, (b) object detections with bounding boxes, (c) Det-LIME explanation heatmap, (d) Det-LIME overlay, (e) LayerCAM, (f) vanilla LIME, (g) LIME variant SLIME, (h) LIME variant DLIME. In the attribution maps, a color spectrum represents relative importance, with warm colors (e.g., red, orange, yellow) indicating high contribution and cool colors (e.g., blue) indicating little or no contribution.}
    \label{fig:lime_comparison_fp_seal}
\end{figure*}

\begin{figure*}[htbp]
    \centering
    \includegraphics[width=0.9\textwidth]{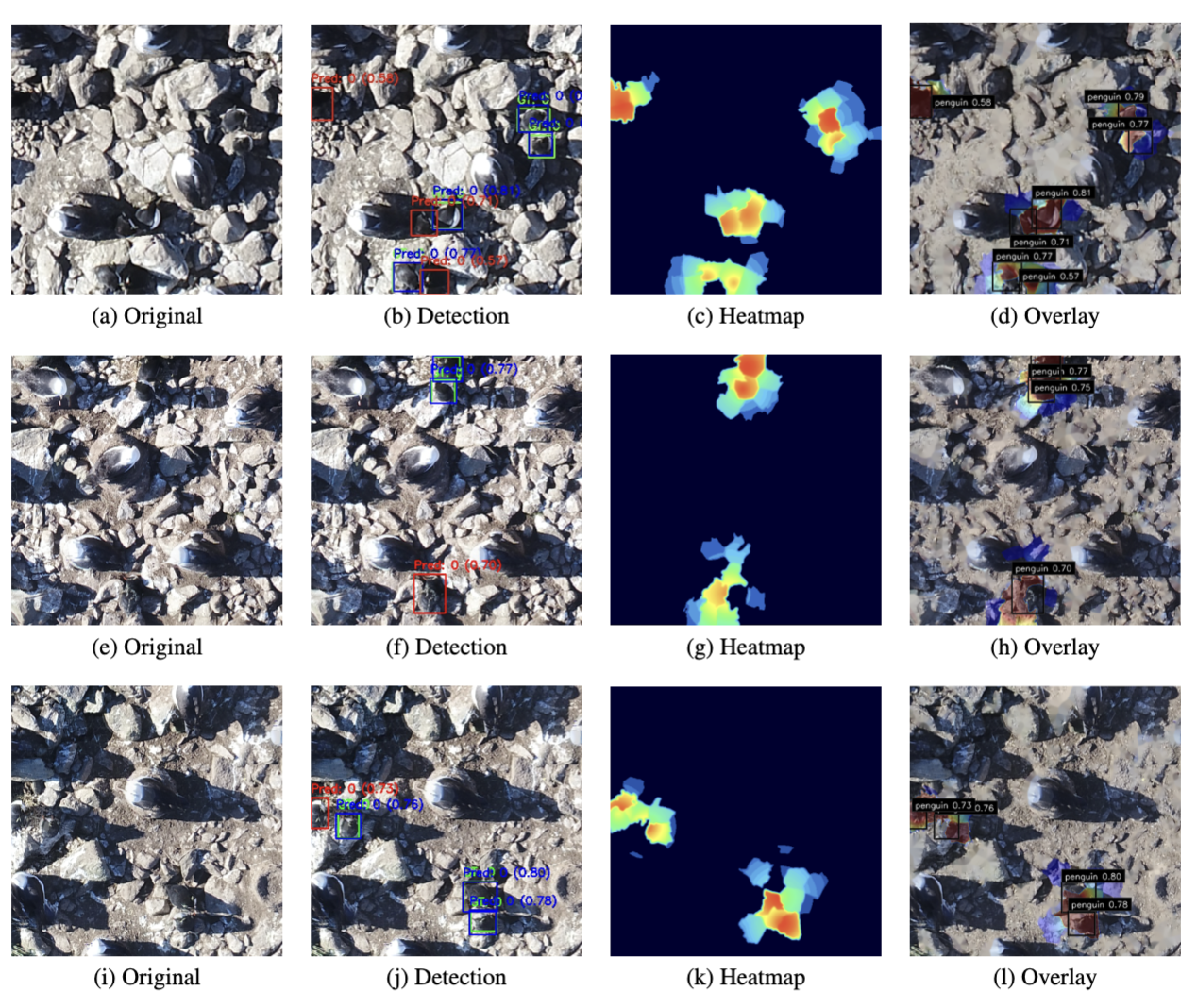}
    \caption{Visual explanations of penguin detection failures. Each column corresponds to one image case. The first column  (a, e, i) shows the original images. The second column (b, f, j) illustrates false positive detections, with ground truth boxes in green, true positives in blue, and false positives in red. The third column (c, g, k) shows Det-LIME explanation heatmaps. The fourth column (d, h, l) shows Det-LIME overlays combining attribution maps with the original images. These visualizations illustrate how Det-LIME identifies image regions that the model focused on when making false positive predictions. Across cases, these errors were frequently located at boundaries, either at the image edge or at transitions between textures such as rock and grass.}
    \label{fig:lime_comparison_fp_penguin}
\end{figure*}

Applying Det-LIME to ecological imagery demonstrated how instance-aware explanations revealed both valid cues and recurrent sources of error (Figure 3 and Figure 4). For correctly identified seals, Det-LIME assigned higher attribution to superpixels overlapping the detected object or its immediate boundary, indicating that the detector relied on image regions associated with the target rather than background context alone. By contrast, vanilla LIME and other LIME variants tended to highlight coarse superpixels spanning both the animal and adjacent background, obscuring the distinction between object and context. This distinction underscored the added value of Det-LIME: while bounding boxes alone indicate where detections occur, attribution maps can help identify the visual evidence supporting those detections and assess whether the detector is relying on animal features, contextual cues, or both. 

Failure cases further illustrated the diagnostic value of Det-LIME. In Figure 3, the region of black ice was incorrectly detected as a seal by Faster R-CNN. Although the false positive bounding box identified the location of the error, it did not reveal the visual evidence underlying the prediction. Det-LIME attribution maps showed concentrated importance over dark, high-contrast textures within the detected region, suggesting that the model relied on low-level visual similarities between seal bodies and surface cracks or melt patterns in the ice.  This instance-level explanation provided evidence for why the false detection may have occurred by showing which image regions most influenced the detector’s output, rather than only indicating the location of the false-positive bounding box. In contrast, LayerCAM produced diffuse activations that extended beyond the detected instance, while vanilla LIME and SLIME provided limited insight into false-positive regions, which reduced their effectiveness in multi-instance detection settings. A similar pattern appeared in penguin detection failures (Figure 4), where Det-LIME explained false positives arising from co-occurring albatrosses and detections near image boundaries, including frame edges and transitions between habitats such as rock and grass. By conditioning explanations on individual detections, Det-LIME revealed the specific visual cues responsible for false positives and supported a more precise analysis of model failure modes than bounding boxes alone.

These examples also show that Det-LIME heatmaps should be interpreted as evidence of model behavior, rather than as direct maps of animal anatomy. In some cases, such as the leftmost seal in Figure 3, the strongest Det-LIME response was located on nearby ice or high-contrast background regions within or close to the detection box, rather than clearly on the seal body. This does not necessarily mean that the background region alone drives the detection. Convolutional neural networks build predictions by combining visual features across the image. Earlier layers typically respond to relatively simple features such as edges, contrast, and texture, while deeper layers combine these features into more complex representations of objects and their surrounding context. A seal detection may therefore depend on both features of the animal and how those features occur in relation to nearby ice, water, shadows, or other parts of the scene. 

Det-LIME does not directly examine these internal representations. Instead, it perturbs different combinations of superpixels and measures how those changes affect a particular detection. High attribution to a background region therefore indicates that changing that region influences the detector's prediction, potentially because the detector uses that information together with features of the animal. Because Det-LIME summarizes these effects as contributions from individual superpixels, relationships among regions are not shown directly in the heatmap. 

For ecological applications, this distinction is important. A heatmap that is not centered on the animal does not necessarily indicate a failure of the explanation method. It may instead reveal that the model is relying on surrounding habitat or contextual information, potentially in combination with animal features, rather than on the animal itself alone. Det-LIME is therefore most useful as a diagnostic tool for identifying whether a detection is supported by animal features, surrounding habitat features, or a mixture of both. 

LayerCAM provided a useful comparison for interpreting these patterns. In Figures 2 and 3, LayerCAM sometimes produced sharper relevance maps that were more closely aligned with visible animal structures than the corresponding Det-LIME maps. This visual clarity can be reassuring when the goal is to confirm that a detection is supported by features of the animal itself, such as body shape, edge structure, or local texture. LayerCAM can provide this type of localization because it uses intermediate feature activations and gradients from within the neural network, allowing it to highlight image regions that are strongly associated with the detected class.

Det-LIME provides a different and complementary view of model behavior. Rather than using internal feature activations, Det-LIME evaluates how changes to image regions affect the detector’s output. This makes it useful for identifying cases where a detection may depend not only on the animal, but also on nearby habitat features, image boundaries, or high-contrast background textures. For example, when Det-LIME assigns strong importance to ice, rock, or habitat edges near a detected animal, this suggests that the detector may be using contextual information that is correlated with the target species in the training data. In ecological applications, this distinction is important because a model can make a correct detection while still relying partly on environmental cues that may not generalize well to new locations, seasons, or image conditions.

This difference reflects a practical trade-off between the two approaches. LayerCAM requires access to the model architecture, internal feature maps, and gradients, which may not be available when researchers use third-party models or deployed detection systems that only return bounding boxes and confidence scores. LayerCAM is also sensitive to the choice of target layer. Lower layers may emphasize edges, textures, or local patterns, while deeper layers may capture broader semantic information related to the animal or its surrounding context. In our examples, the explanation quality varied depending on which layer was selected, indicating that layer choice can affect how clearly the resulting map supports ecological interpretation. Det-LIME is less visually fine-grained, but because it only requires the input image and detector outputs, it can be applied more broadly as a black-box diagnostic tool for evaluating whether detections are driven by animal features, contextual features, or a mixture of both. When model internals are available, LayerCAM and Det-LIME should be viewed as complementary rather than competing approaches. Used together, they provide a more diverse and comprehensive perspective on model behavior, helping researchers distinguish detections supported by visible animal structures from those influenced by habitat context or background cues.

The results of the present study illustrate that Det-LIME adapts the perturbation-based framework of LIME to ecological object detection in a robust manner, producing explanations that are box-conditioned and instance-stable.  Perturbations were generated via superpixel masking, and detections were matched back to original instances using IoU-based correspondence across perturbed outputs. Instance contributions were weighted by detection confidence and spatial alignment, with only sufficiently overlapping detections retained via an IoU threshold. This improved the stability of instance-level attribution across perturbations, reduced attribution drift, and suppressed weakly aligned background responses. We demonstrated instance fidelity using the Attribution Ratio and Max Saliency Hit Rate, and showed that Det-LIME outperformed vanilla LIME and LIME variants SLIME and DLIME for multi-instance detection tasks by keeping saliency concentrated within the queried detection and consistently anchoring peak attribution on the same instance.

A key contribution of Det-LIME lies in its diagnostic utility for multiple predictions within single images, a frequently encountered problem for researchers using “black box” AI approaches to study colonial or socially aggregating organisms. Explanations revealed that the seal detectors often misattributed dark background regions, such as black ice, leading to false positives when low-level visual similarity overwhelmed biological distinctiveness. In the penguin example, the detector misclassified albatrosses or produced false detections along habitat boundaries. These insights suggest concrete avenues for model improvement: targeted data augmentation with negative examples of ice and rock and refined training sets with greater habitat variability. Beyond improving models, attribution maps signal when outputs should be treated cautiously, offering a safeguard against over-reliance on automated predictions in conservation workflows.

At the same time, several limitations warrant consideration. Det-LIME inherits LIME’s reliance on superpixels and perturbation kernels, which can introduce sensitivity to segmentation parameters and discretize importance into regions rather than addressing pixel-level fidelity. Our evaluation also measured localization fidelity against bounding boxes, which may understate explanation quality when salient regions align with subparts of an animal or spill beyond annotated boundaries. Residual labeling errors further complicate interpretation, particularly in ecological datasets where annotations are resource-intensive and often noisy. 

In addition, the present study evaluated Det-LIME as a complete workflow rather than isolating the contribution of each individual component. The components of Det-LIME are sequentially connected, with each step relying on the output of the previous one, and the intermediate results are not independent attribution maps that can be evaluated in the same way as the final explanation. Consequently, the reported metrics assess the performance of the full explanation pipeline rather than that of individual components. A systematic ablation study would help quantify the contribution of each stage and remains an important direction for future work. 

The proposed evaluation metrics, AR and MSHR, also have inherent limitations. Because object-detection explanations are generated with respect to model-predicted bounding boxes, class labels, and confidence scores, these metrics evaluate how well an explanation aligns with the detector's outputs rather than providing a fully detector-independent measure of explanation quality. This limitation is not unique to Det-LIME, but reflects a broader challenge in explainable artificial intelligence (XAI). Previous studies have noted that interpretability lacks a universally accepted definition or evaluation standard, and that reliable ground-truth explanations are rarely available in real-world applications \citep{doshivelez2017rigorous}. As a result, XAI methods are typically assessed using complementary proxy measures, such as localization, faithfulness, robustness, or human evaluation, each capturing a different aspect of explanation quality \citep{nauta2023anecdotal}. Future work could incorporate perturbation-based faithfulness evaluations, including deletion and insertion analyses, to examine whether regions highlighted by Det-LIME have a measurable effect on detector confidence or localization. Such analyses would complement the spatial overlap metrics presented in this study by providing additional evidence that the highlighted regions meaningfully influence model predictions. Finally, our study did not assess runtime efficiency, operator burden, or field usability, which are critical factors for deployment and should be explored in future work.

Despite these challenges, Det-LIME demonstrated how explanation methods can serve as actionable diagnostics for both ecological inference and model development. By revealing the visual evidence underpinning predictions, Det-LIME helped build justified confidence in correct detections while also highlighting recurrent weaknesses that can guide ecologists toward targeted data collection, model refinement, and field validation. More broadly, these results underscore the role of explainability as a bridge between computer vision models and ecological decision-making, ensuring that detectors are not only accurate but also evidence-based, clear about where they fail, and defensible in policy and field workflows.

\section*{Note on references}
In computer science and machine learning, many high-impact and foundational contributions are published in double-blind peer-reviewed conference and workshop proceedings rather than traditional journals. Accordingly, several key references cited in this work appear in conference proceedings that serve as primary archival venues within these fields.

\bibliographystyle{plainnat}
\bibliography{preprint}

\appendix
\section{Data Collection and Labeling}
\subsection{Harbor Seal Data}
Drone surveys were conducted using a Wingtra One Gen II fixed-wing platform (Zurich, Switzerland) with vertical takeoff and landing (VTOL) capabilities imaging with a Sony Alpha 6100 APS-C camera (Tokyo, Japan) with a Sony E $20 \mathrm{~mm} \mathrm{f} / 2.8$ lens. Flight plans were created and carried out using WingtraPilot flight planning software. Flights operated at $\sim 60-85 \mathrm{~m}$ altitude and $\sim 9-22 \mathrm{~m} / \mathrm{s}$ airspeed over regions that were historically sampled by occupied aircrafts. Drone operations were conducted under permit by NOAA and the NPS.

Flights in glacial fjords occurred along non-overlapping parallel transects oriented lengthwise through the glacial end of the fjord, with transects oriented perpendicular to the glacier terminus. These flight plans were like those of historic surveys that used occupied aircrafts \citep{womble2020calibrating}, but were optimized to achieve high-density coverage of the inner regions of the fjords where seal densities are highest. Surveys from occupied aircrafts historically sampled the entire west arm of JHI along the same 12 transects year after year, maintaining a $\sim 100-\mathrm{m}$ buffer between photographs across transects and a $\sim 20-\mathrm{m}$ buffer between consecutive photographs along transects. Surveys from unoccupied aircrafts in 2023 and 2024 surveyed smaller gross extents with a series of nearly contiguous but not overlapping transects, maintaining a $\sim 5-\mathrm{m}$ buffer between photographs across transects and 65-70
\% overlap between consecutive photographs along transects. Surveys from unoccupied aircrafts in glacial fjords consisted of 1-3 flights each using impromptu flight plans informed by the extent of floating ice habitat and drone performance in the prevailing weather conditions at the time of the survey. Surveys of terrestrial sites also consisted of parallel transects arranged in a high density to achieve a target of $\sim 601 \%$ overlap between photographs along transects and across transects.

The training dataset was visually reviewed and manually thinned to remove photographs that did not include at least one positive instance of a harbor seal on floating ice. This was done to mitigate the risk of negative bias in model training, which can occur with an overabundance of negative training data. The resulting dataset images were each subdivided into tiles of $640 \times 640$ pixels. Each tile was manually inspected and annotated to mark all harbor seal locations using LabelMe \citep{russell2008labelme}.

\section{Modeling Method}
\subsection{Harbor Seal Detection Task using Faster R-CNN}
We employed a transfer learning approach for seal detection. Our model, a Faster R-CNN \citep{ren2016fasterrcnnrealtimeobject} with a ResNet-50 backbone and Feature Pyramid Network (FPN), was initialized with weights pre-trained on the COCO dataset. We then performed full fine-tuning, updating all layers of the network to adapt the model to our seal dataset.

To improve robustness, the training dataset was augmented with geometric transformations (horizontal and vertical flips, rotations up to $45^{\circ}$, and random crops) and color adjustments (brightness, contrast, saturation, and hue). Validation images were processed only with normalization and tensor conversion for consistency, while the test set was completely held out and used only for final evaluation without any data augmentation.

The model was trained using stochastic gradient descent with gradient clipping, warmup learning rate scheduling, and early stopping. Weights \& Biases was used for experiment tracking.

Hyperparameters were tuned over learning rate (0.001-0.01), momentum (0.85-0.95), and weight decay (0.0001-0.001) using a grid search. The chosen configuration used a batch size of 8, a learning rate of 0.0090, momentum of 0.874, and weight decay of 0.0001. Although training was set for up to 200 epochs, the best model was obtained at epoch 67, where validation performance peaked with a mean Average Precision (mAP, averaged over IoU thresholds from 0.5 to 0.95) of 0.61, mAP50 of 0.95, and mean Average Recall (mAR, recall averaged across up to 100 detections per image) of 0.69. The corresponding training and validation losses were 0.11 and 0.13, respectively. On the held-out test set of 76 images, the final model achieved an overall mAP of 0.65, with mAP50 of 0.98 and mAP75 of 0.78.

\subsection{Penguin Detection Task using YOLOv9
}
We trained an object detection model to identify penguins using the YOLOv9c architecture \citep{wang2024yolov9learningwantlearn}. The dataset contained 2,644 training images, 288 validation images, and 321 test images, all reformatted into the YOLOv9 format for consistency. To improve generalization and reduce overfitting, we applied a range of data augmentation techniques, including random rotation, translation, scaling, horizontal flips, and color space adjustments, along with mixup regularization. Training was carried out on 640-pixel images with a batch size of 16 for up to 200 epochs, with early stopping if validation performance did not improve for 30 epochs. We optimized the model with AdamW, using an initial learning rate of 0.01, momentum of 0.937, and weight decay of 0.0005. On the validation set, the model achieved robust performance, with a precision of 0.911, a recall of 0.923, an mAP50 of 0.956, and an mAP50–95 of 0.519. Evaluation on the held-out test set confirmed this robustness, yielding a precision of 0.912, a recall of 0.933, an mAP50 of 0.960, and an mAP50–95 of 0.528 across 2,720 instances. These results show that YOLOv9c can effectively detect penguins in challenging natural imagery and generalize well to unseen data.

\end{document}